\documentclass{wscpaperproc}
\usepackage{latexsym}
\usepackage{graphicx}
\usepackage{mathptmx}
\usepackage[T1]{fontenc}
\usepackage{wrapfig}
\usepackage{caption} 
\usepackage{wrapfig,booktabs}

\usepackage{amsmath}
\usepackage{amsfonts}
\usepackage{amssymb}
\usepackage{amsbsy}
\usepackage{amsthm}
\usepackage{algorithm}
\usepackage{algpseudocode}
\makeatletter
\newcommand\floatc@ruledwsc[2]{%
  \setbox\@tempboxa\hbox{#1: #2}%
  \ifdim\wd\@tempboxa>\hsize
    #1: #2\par
  \else
    \hb@xt@\hsize{\hfil\box\@tempboxa\hfil}%
  \fi}
\newcommand\fs@ruledwsc{%
  \def\@fs@cfont{}\let\@fs@capt\floatc@ruledwsc
  \def\@fs@pre{\hrule height.8pt depth0pt \kern2pt}%
  \def\@fs@post{\kern2pt\hrule\relax}%
  \def\@fs@mid{\kern2pt\hrule\kern2pt}%
  \let\@fs@iftopcapt\iftrue}
\makeatother
\floatstyle{ruledwsc}
\restylefloat{algorithm}

\usepackage[pdftex,colorlinks=true,urlcolor=blue,citecolor=black,anchorcolor=black,linkcolor=black]{hyperref}

\newtheoremstyle{wsc}
{3pt}
{3pt}
{}
{}
{\bf}
{}
{.5em}
{}

\theoremstyle{wsc}

\begin{document}

%
%

\pagestyle{fancyplain}

\thispagestyle{plain}
\firstPageHead{}

\chead{\fancyplain{}{\itshape Salki\'{c}, Fichtl, Ulrich, Ehm, Bonik, and Groh}}

\rhead{}
\cfoot{}
\renewcommand{\headrulewidth}{0pt} 

\makeatletter
\let\@internalcite\cite
\def\cite{\def\@citeseppen{-1000}%
    \def\@cite##1##2{(##1\if@tempswa , ##2\fi)}%
    \def\citeauthoryear##1##2##3{##2 ##3}\@internalcite}
\def\citeNP{\def\@citeseppen{-1000}%
    \def\@cite##1##2{##1\if@tempswa , ##2\fi}%
    \def\citeauthoryear##1##2##3{##2 ##3}\@internalcite}
\def\citeN{\def\@citeseppen{-1000}%
    \def\@cite##1##2{##1\if@tempswa, ##2)\else{}\fi}%
    \def\citeauthoryear##1##2##3{##2 (##3)}\@citedata}
\def\citeA{\def\@citeseppen{-1000}%
    \def\@cite##1##2{(##1\if@tempswa , ##2\fi)}%
    \def\citeauthoryear##1##2##3{##1}\@internalcite}
\def\citeANP{\def\@citeseppen{-1000}%
    \def\@cite##1##2{##1\if@tempswa , ##2\fi}%
    \def\citeauthoryear##1##2##3{##1}\@internalcite}
\def\shortcite{\def\@citeseppen{-1000}%
    \def\@cite##1##2{(##1\if@tempswa , ##2\fi)}%
    \def\citeauthoryear##1##2##3{##2 ##3}\@internalcite}
\def\shortciteNP{\def\@citeseppen{-1000}%
    \def\@cite##1##2{##1\if@tempswa , ##2\fi}%
    \def\citeauthoryear##1##2##3{##2 ##3}\@internalcite}
\def\shortciteN{\def\@citeseppen{-1000}%
    \def\@cite##1##2{##1\if@tempswa, ##2\else{}\fi}%
    \def\citeauthoryear##1##2##3{##2 (##3)}\@citedata}
\def\shortciteA{\def\@citeseppen{-1000}%
    \def\@cite##1##2{(##1\if@tempswa , ##2\fi)}%
    \def\citeauthoryear##1##2##3{##2}\@internalcite}
\def\shortciteANP{\def\@citeseppen{-1000}%
    \def\@cite##1##2{##1\if@tempswa , ##2\fi}%
    \def\citeauthoryear##1##2##3{##2}\@internalcite}
\def\citeyear{\def\@citeseppen{-1000}%
    \def\@cite##1##2{(##1\if@tempswa , ##2\fi)}%
    \def\citeauthoryear##1##2##3{##3}\@citedata}
\def\citeyearNP{\def\@citeseppen{-1000}%
    \def\@cite##1##2{##1\if@tempswa , ##2\fi}%
    \def\citeauthoryear##1##2##3{##3}\@citedata}
%
%
%
\def\@citedata{%
    \@ifnextchar [{\@tempswatrue\@citedatax}%
                  {\@tempswafalse\@citedatax[]}%
}

\def\@citedatax[#1]#2{%
\if@filesw\immediate\write\@auxout{\string\citation{#2}}\fi%
  \def\@citea{}\@cite{\@for\@citeb:=#2\do%
    {\@citea\def\@citea{, }\@ifundefined
       {b@\@citeb}{{\bf ?}%
       \@warning{Citation `\@citeb' on page \thepage \space undefined}}%
{\csname b@\@citeb\endcsname}}}{#1}}%

%
\def\@citex[#1]#2{%
\if@filesw\immediate\write\@auxout{\string\citation{#2}}\fi%
  \def\@citea{}\@cite{\@for\@citeb:=#2\do%
    {\@citea\def\@citea{; }\@ifundefined
       {b@\@citeb}{{\bf ?}%
       \@warning{Citation `\@citeb' on page \thepage \space undefined}}%
{\csname b@\@citeb\endcsname}}}{#1}}%

%
\def\@biblabel#1{}
\makeatother



\newdimen\bibindent
\bibindent=0.0em
\def\thebibliography#1{\section*{\refname}\list
   {}{\settowidth\labelwidth{[#1]}
   \leftmargin\parindent
   \itemindent -\parindent
   \listparindent \itemindent
   \itemsep 0pt
   \parsep 0pt}
   \def\newblock{}
   \sloppy
   \sfcode`\.=1000\relax}


\setlength{\baselineskip}{12.7pt}

\title{A SYSTEMATIC APPROACH TO CONSTRUCTING A CHANCE-AND-RISK MATRIX FOR SEMICONDUCTOR SUPPLY CHAINS}

\author{\begin{center}Ema Salki\'{c}\textsuperscript{1,2}, Alexander Fichtl\textsuperscript{2}, Philipp Ulrich\textsuperscript{1}, Hans Ehm\textsuperscript{1}, Marta Bonik\textsuperscript{1}, and Georg
Groh\textsuperscript{2}\\
[11pt]
\textsuperscript{1}Infineon Technologies AG, Munich, GERMANY\\
\textsuperscript{2}TUM School of Computation, Information and Technology, Technical University of Munich, Munich, GERMANY
\end{center}
}

\maketitle

\vspace{-12pt}

\section*{ABSTRACT}
Semiconductor supply chains face escalating risks from geopolitical tensions, geographic concentration, and rapid technological shifts, yet no scalable system continuously extracts, structures, and prioritizes risk intelligence from public corporate disclosures.
We present an end-to-end pipeline that retrieves corporate documents for semiconductor companies and uses large language models (LLMs) to extract the risks and opportunities they describe.
It organizes these into a knowledge graph linking each item to its category, sources, and related events, then merges duplicates and ranks them with a three-layer mechanism combining an algorithmic formula, an LLM relevance adjustment, and expert validation.
Applied to five companies across the value chain, the pipeline produces 76{,}207 scored items, of which an independent check finds 92.6\% valid.
The automated rankings match expert judgment at an average Spearman correlation of 0.55 for risks and 0.72 for opportunities, and the resulting matrices identify trade restrictions as the dominant cross-company risk.

\section{INTRODUCTION}
\label{sec:intro}

Semiconductors are fundamental to almost every modern industry, with the global market reaching about USD~500~billion in 2024 \cite{sia2024factbook}.
However, the supply chains that produce them are structurally fragile. Front-end fabrication cycles span months, a single advanced fab requires over USD~10~billion in capital, and multi-tier dependencies concentrate critical capacity in a few regions \cite{kleinhans2020global,acemoglu2012network}.
The 2020--2023 global chip shortage, which halted automotive production lines worldwide and imposed GDP losses estimated in the hundreds of billions of dollars, laid bare these vulnerabilities \cite{simchi2022test,ivanov2014ripple}.
Escalating US--China export controls have since added a persistent layer of geopolitical uncertainty \cite{ivanov2021digital}.

Governments have responded with large-scale industrial policy.
The European Union adopted the European Chips Act in 2023, mobilizing up to EUR~43~billion to double Europe's share of global chip production to 20\% by 2030 \cite{eu_chips_act2023}. The United States committed USD~52.7~billion through its CHIPS and Science Act \cite{us_chips_act2022}.
However, high-profile project cancellations, such as Intel's EUR~30~billion Magdeburg fab and Wolfspeed's Ensdorf facility, suggest that financial incentives alone are insufficient \cite{eca2025chips}.
What Europe additionally needs is better intelligence and the ability to continuously monitor supply chain risks, detect emerging opportunities, and act before disruptions materialize.

\begin{figure}[ht]
\centering
\includegraphics[width=0.65\columnwidth]{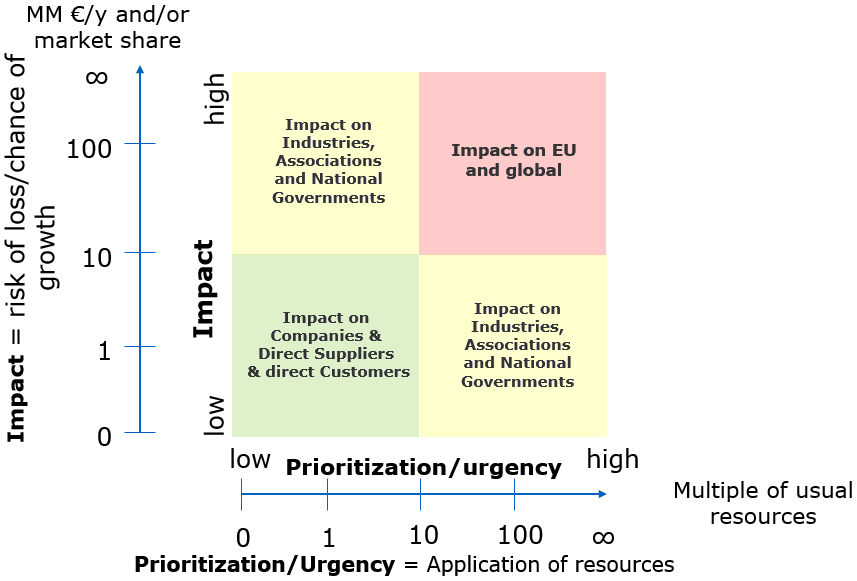}
\caption{ALLPROS chance-and-risk matrix. The vertical axis measures estimated impact in millions of euros; the horizontal axis measures the resources required to address each item.}
\label{fig:chance-risk-matrix}
\end{figure}

One institutional response is the Industrial Alliance on Processors and Semiconductor Technologies (ALLPROS), established under Pillar~III of the European Chips Act as a coordination mechanism that brings together semiconductor companies, EU Member State governments, and the European Commission \cite{eu_chips_act2023,allpros2024}.
ALLPROS organizes its work into four working groups, including Supply Chains, Skills, Per- and Polyfluoroalkyl Substances (PFAS), and Automotive Electronics, and maps the European semiconductor value chain across twelve clusters, ranging from raw materials and equipment to electronics manufacturing services and end-user applications.
A cornerstone deliverable of the Supply Chain working group is the \emph{chance-and-risk matrix}. This prioritization tool plots identified risks and opportunities along two axes, namely estimated impact in millions of euros and the resources required to address them. Items in the upper-right quadrant represent the most critical issues requiring coordinated European-level action (Figure~\ref{fig:chance-risk-matrix}).

The current Alliance workflow for adding to the matrix combines AI-generated metrics with human expert review. However, no scalable, automated system exists to continuously feed the matrix with structured, scored intelligence derived from publicly available corporate documents.
This paper presents an end-to-end pipeline that addresses this gap. Our system retrieves corporate documents for semiconductor companies, extracts risk and opportunity instances using large language models (LLMs), consolidates them into a knowledge graph (KG) based on the Web Ontology Language (OWL) via semantic clustering, and scores them using a three-layer mechanism that combines algorithmic analysis, LLM relevance re-scoring, and domain expert validation.
We apply the pipeline to five target companies spanning the value chain and produce scored rankings that map directly onto the Alliance's chance-and-risk matrix.

Unlike existing approaches that either automate data collection without structuring or provide structured ontologies without automation, our pipeline uniquely combines both by transforming unstructured public documents into machine-queryable, scored, ontology-backed intelligence that replaces weeks of manual analyst work with a repeatable, auditable process.
The pipeline is also modular, so an analyst can substitute the automated web crawl with a manually curated document set (e.g., specific supplier filings or policy documents), executing only the downstream stages for targeted due diligence or rapid policy-impact assessments.

\section{BACKGROUND}
\label{sec:background}

Our pipeline rests on the premise that risks and opportunities are two manifestations of the same underlying uncertainty \cite{hillson2002extending}, a view codified by ISO~31000 \cite{iso31000}.
Following ISO~31000, we treat a risk as the effect of uncertainty on objectives when that effect is potentially harmful, and an opportunity as the same effect when it is potentially beneficial.
We review the three research threads it draws on: supply chain risk management (SCRM), ontology-based knowledge representation, and LLM-based extraction, and then position our contribution relative to existing work.

\subsection{Supply Chain Risk Management in Semiconductors}

SCRM addresses the identification and management of risks within and external to a supply chain \cite{juttner2003supply}.
These challenges are especially acute in semiconductors, where fabrication is geographically concentrated, capital-intensive, and spans hundreds of steps over several months \cite{kleinhans2020global}.
These characteristics have driven growing interest in automated approaches that continuously structure risk-relevant information from public sources \cite{ivanov2021digital}.

\subsection{Ontologies, Knowledge Graphs (KGs), and LLM-Based Extraction}

An \emph{ontology} provides a formal vocabulary of classes, properties, and constraints for a domain \cite{gruber1993translation}, while a \emph{KG} instantiates this schema with concrete entities and typed relationships \cite{hogan2021knowledge}.
We use OWL \cite{bernerslee2001semantic} as the representation language, as it provides the most expressive layer of the Semantic Web stack. Beyond simple class hierarchies, it enables shared, machine-readable domain vocabularies and captures complex relationships via disjointness axioms, typed object properties with explicit domain and range constraints, and inverse properties, all of which are central to our ontology design.

Recent advances in LLMs have made \emph{in-context learning} \cite{brown2020language} viable, enabling zero-shot classification \cite{yin2019benchmarking} and structured extraction via taxonomy-guided prompts without task-specific training. When extraction operates at scale, Sentence-BERT \cite{reimers2019sentencebert}, a sentence embedding variant of BERT (Bidirectional Encoder Representations from Transformers), enables semantic deduplication by clustering records whose cosine similarity exceeds a threshold, and the LLM-as-a-judge paradigm \cite{zheng2023judging} allows models to assess and score extracted content.

\subsection{Positioning Relative to Prior Work}

Recent SCRM literature reviews \cite{emrouznejad2023supply,choudhary2022risk,ho2015supply} have identified a growing need for automated, data-driven risk identification at scale, noting that the dominant assessment techniques such as the Analytic Hierarchy Process (AHP), Failure Mode and Effects Analysis (FMEA), and Bayesian Networks remain dependent on manual expert input and periodic reassessment.

On the natural language processing (NLP) and LLM side, \citeN{zhao2024optimizing} proposed an LLM-based framework for supply chain risk identification and categorization, demonstrating that LLMs can automate risk extraction from diverse unstructured sources; however, their framework does not store extracted risks in a structured knowledge representation and provides no scoring or ranking mechanism.

A complementary thread has explored the use of KGs for risk intelligence. \citeN{mahfouz2021framework} built a KG system at J.P.~Morgan that matches institutional risks to multilingual news articles using neural embeddings, achieving 96.6\% accuracy in expert evaluation. However, its KG is built ad hoc rather than from a formal ontology, limiting the extensibility and reusability needed to consolidate heterogeneous views across alliance actors, and it does not score or rank risks. \citeN{li2024survey} surveyed 246 articles on KG applications in enterprise risk management and identified persistent challenges, including scarcity of domain-specific training data and limited real-time capability.
Most closely related to our architecture, \citeN{zheng2025empowering} proposed an ontology-guided method for extracting supply chain risk KGs from news articles using LLMs, producing a KG with over 30{,}000 entities; however, their work addresses natural hazards rather than semiconductors, does not model opportunities, and does not score or rank the extracted risks.

Within the semiconductor domain, \citeN{razouk2022ai} combine KG embeddings with BERT-based causal extraction from FMEA documents for root cause analysis, and \citeN{kuo2023semantic} develop a semantic-web-based risk assessment framework for collaborative supply chain planning.
Both operate on internal enterprise data rather than external corporate disclosures and focus on a single risk dimension rather than the dual risk-and-opportunity perspective. Our work addresses these limitations within an active institutional context (ALLPROS), making the outputs actionable policy deliverables rather than academic artifacts.

\section{METHODOLOGY}
\label{sec:methodology}

\begin{figure}[h]
\centering
\includegraphics[width=\columnwidth]{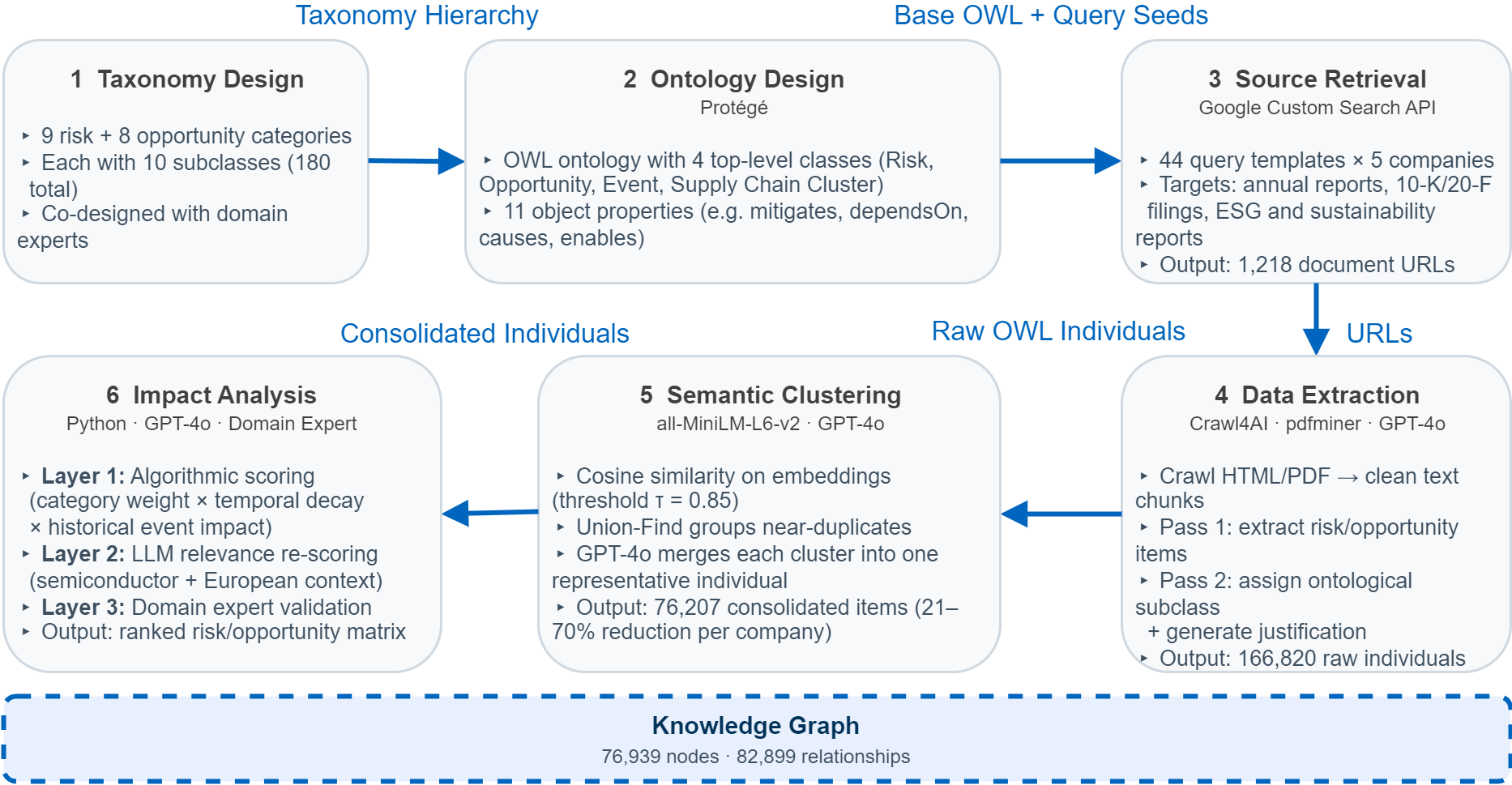}
\caption{End-to-end pipeline overview. Stages~1--3 (top row, left to right) produce the ontology schema and retrieve source documents. Stages~4--6 (bottom row, right to left) extract, cluster, and score risks and opportunities.}
\label{fig:pipeline-overview}
\end{figure}

We propose an end-to-end pipeline that constructs a dynamic KG capturing supply chain risks and opportunities for semiconductor companies.
The pipeline consists of six stages. (1)~\emph{Taxonomy definition} builds a two-level hierarchy of risk and opportunity categories. (2)~\emph{Ontology design} formalizes that hierarchy in OWL and adds the typed relationships and constraints a flat taxonomy cannot express. (3)~\emph{Source retrieval} collects public corporate documents. (4)~\emph{Data extraction} uses the LLM to pull individual risks and opportunities from them. (5)~\emph{Semantic clustering and consolidation} merges items that describe the same concept. (6)~\emph{Impact analysis} scores and ranks the results.
All LLM-based stages use OpenAI's GPT-4o (snapshot \texttt{gpt-4o-2024-08-06}, accessed via the OpenAI API between January and March 2025) \cite{openai2024gpt4o} with temperature 0.0 (greedy decoding) to maximize output consistency across runs.
We apply the pipeline to five target companies, including Intel, Infineon Technologies, Texas Instruments, Air Liquide, and Siltronic, all of which are members of the ALLPROS alliance \cite{allpros2024}.
These companies were selected to span distinct value-chain segments (integrated device manufacturing, fabless design, specialty chemicals, silicon wafer manufacturing) and to ensure domain experts at an ALLPROS member company were available to validate the resulting rankings for each segment.
Figure~\ref{fig:pipeline-overview} provides an overview of the six-stage pipeline with its inputs and outputs.

\subsection{Ontology Design}

The backbone of the pipeline is an OWL ontology that goes well beyond a simple hierarchical classification.
We adopt an ontology-backed KG rather than flat files or a relational database because it lets the continuously growing instance set gain new typed relationships without schema migration, enforces the \emph{Risk}--\emph{Opportunity} disjointness declaratively, and encodes the causal links that let one event both pose a risk and create an opportunity.
While the ontology \emph{includes} taxonomic hierarchies for risks (9~categories, 90~subclasses) and opportunities (8~categories, 90~subclasses), it enriches them with typed object properties, disjointness axioms, and cross-class relationships that a plain taxonomy cannot express.
Eight category pairs are shared across the two hierarchies, reflecting the dual nature of uncertainty \cite{hillson2002extending}; the risk side adds \emph{Reputational Risk}, which has no natural opportunity counterpart.
The hierarchical structures were developed through a four-phase process: (1)~domain research on semiconductor industry reports, (2)~LLM-assisted category generation \cite{nickerson2013taxonomy}, (3)~manual expert curation, and (4)~keyword enrichment with 15--30 domain-specific terms per entry.

\begin{figure}[ht]
\centering
\includegraphics[width=\columnwidth]{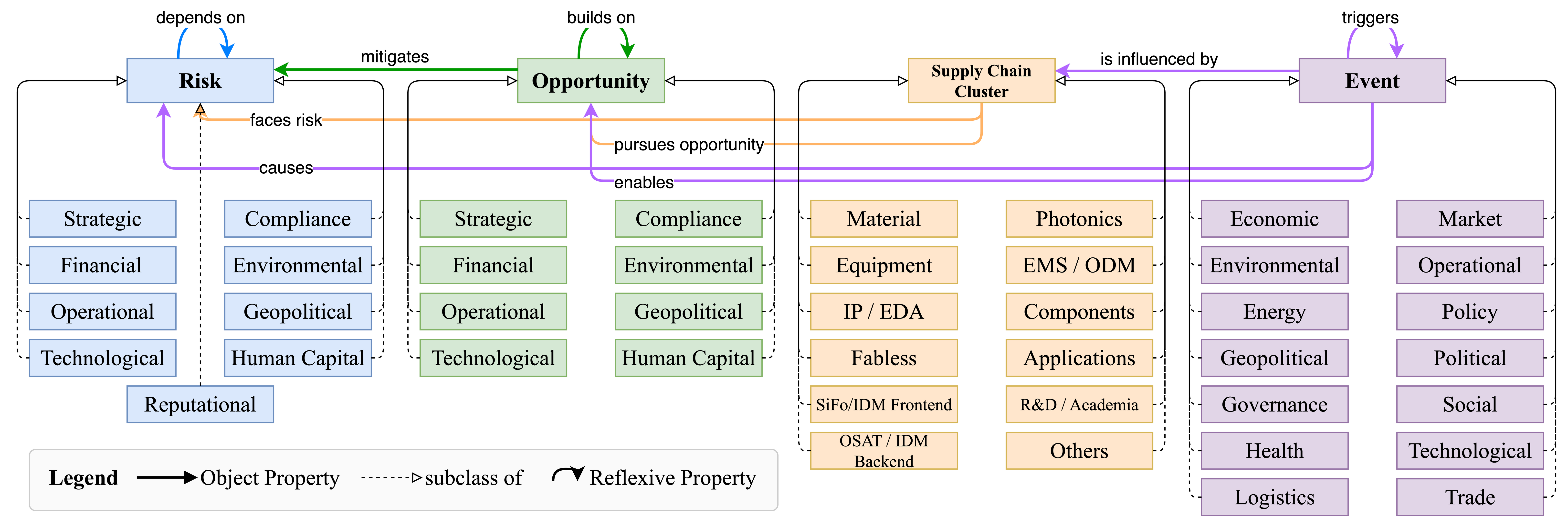}
\setlength{\belowcaptionskip}{-10pt}
\caption{OWL class hierarchy of the four top-level ontology classes and their first-level subclasses. Every subclass contains pre-populated individuals.}
\label{fig:ontology-hierarchy}
\end{figure}

The ontology defines four top-level classes.
\emph{Risk} and \emph{Opportunity} are declared disjoint and encode the taxonomy hierarchies with their data properties (definitions, keywords, synonyms) and three intra-taxonomy object properties, including \emph{depends on} (risk-to-risk cascading), \emph{builds on} (opportunity-to-opportunity reinforcement), and \emph{mitigates} (opportunity-to-risk mitigation with inverse \emph{has mitigation}).
\emph{Supply Chain Cluster} groups 128 companies into 12 functional segments of the semiconductor value chain (e.g., silicon foundries, fabless designers, equipment suppliers) and connects them to risks and opportunities through \emph{faces risk} and \emph{pursues opportunity} properties.
\emph{Event} captures historical occurrences across 14 event types, each carrying magnitude, likelihood, relevance, and duration scores.
Events connect to the rest of the ontology through \emph{causes} (event-to-risk), \emph{enables} (event-to-opportunity), \emph{triggers} (event-to-event), and \emph{is influenced by} (event-to-company) properties.
This four-class structure forms a foundational knowledge layer that provides classification context, causal structure, and historical grounding for the company-specific data extracted in subsequent stages.
Figure~\ref{fig:ontology-hierarchy} shows the complete class hierarchy; every leaf class is pre-populated with curated OWL individuals that seed the KG before any automated extraction begins.

\subsection{Source Retrieval and Data Extraction}

We use company-disclosed documents (annual reports, 10-K filings, environmental, social, and governance (ESG) reports, sustainability reports, investor presentations, governance documents) as the sole input source because they are audited, legally mandated, and publicly accessible.
For each of the five target companies, we define 44 search queries combining the company name, a document type, and a target year (2024 or 2025), and execute them against the Google Custom Search JSON API to retrieve relevant URLs.

The data extraction stage transforms the retrieved documents into structured OWL individuals through four phases.
First, we crawl each URL and extract text content, routing HTML pages through a Markdown converter and PDF files through a dedicated text extractor.
Second, we apply rule-based text cleaning to remove boilerplate (e.g., cookie banners, navigation menus, legal footers) and split long documents into fixed-size chunks of 200 lines each.
Third, we apply a two-pass LLM extraction.
The first pass determines whether a chunk contains any risk or opportunity signal, discarding irrelevant content.
The second pass injects the full taxonomy (all 180 subclasses with their keywords) into the prompt and instructs the model to extract every distinct risk and opportunity, returning for each item a descriptive name, an assigned taxonomy subclass, and a one-sentence justification.
Fourth, we generate OWL instance blocks from the parsed output, converting each name to a PascalCase IRI, normalizing subclass labels through case-insensitive lookup with substring fallback, and performing two-pass deduplication (exact IRI match followed by normalized matching).

Each individual is typed as \emph{Scraped Risk} or \emph{Scraped Opportunity}, a subclass of the foundational \emph{Risk} or \emph{Opportunity} class.
The assigned taxonomy subclass is stored via \emph{has assigned risk subclass} or \emph{has assigned opportunity subclass}, and provenance (source URLs, publication years) is preserved as data properties.

\subsection{Semantic Clustering and Consolidation}

The extraction stage intentionally casts a wide net, producing thousands of individuals per company.
Many of these refer to the same underlying concept expressed differently across documents \cite{elmagarmid2007duplicate}.
The name-based deduplication from the previous stage catches only exact matches, so we apply a deeper deduplication using embedding-based semantic clustering.

We encode each individual's label into a 384-dimensional vector using all-MiniLM-L6-v2, a distilled Sentence-BERT model \cite{reimers2019sentencebert}, and compute pairwise cosine similarities.
Two individuals are placed in the same cluster if their cosine similarity exceeds a threshold $\tau = 0.85$.
We chose this threshold conservatively rather than through a formal sensitivity analysis: lower values merge more aggressively but risk conflating genuinely distinct items, while higher values preserve finer distinctions but leave more near-duplicates unconsolidated. Because the embeddings are cached, a systematic threshold sweep is an inexpensive next step.
We implement transitive closure using the Union-Find data structure with path compression and union-by-rank.
Individuals that do not meet the threshold with any other individual remain as singletons and pass through unchanged.

Before consolidation, we assess each cluster's internal consistency using the \emph{agreement ratio} $\alpha$, defined as the proportion of members sharing the majority taxonomy subclass.
\emph{Perfect clusters} ($\alpha = 1.0$) carry a single confirmed subclass into consolidation; \emph{non-perfect clusters} ($\alpha < 1.0$) preserve all original subclass assignments.
The agreement ratio is a descriptive measure rather than a tuned parameter, and a value of 1.0 simply marks clusters in which all members share the same subclass.
For each cluster, GPT-4o receives all member names and justifications and returns a consolidated name and merged justification, while all source references and publication years are aggregated to maintain provenance.

\subsection{Impact Analysis and Expert Validation}

Even after consolidation, each company retains hundreds of items.
We rank them using a two-level scoring pipeline to surface the most critical risks and promising opportunities.
The first level computes a deterministic impact score for each individual as the product of three factors, $S = w_c \cdot w_t \cdot f_e$.

The \emph{category weight} $w_c$ reflects the systemic importance of the parent taxonomy category on a four-tier scale (Systemic~1.4, High~1.2, Medium~1.0, Low~0.8), set in consultation with domain experts.
Categories whose disruptions cascade across supply chain tiers (Geopolitical, Compliance, Environmental) receive higher weights, while more localized categories (Reputational, Human Capital) receive lower weights.

The \emph{temporal weight} $w_t = \exp(-\lambda \cdot (T - t_{\max}))$ applies exponential decay based on the most recent publication year $t_{\max}$ among the individual's source documents, with decay rate $\lambda = 0.1$ and reference year $T = 2025$.
Items with no publication year receive a default weight of 0.5.

The \emph{event factor} $f_e = 1 + \beta \cdot \ln\!\bigl(1 + \sum_{e \in E} \mathrm{ImpactScore}(e)\bigr)$ amplifies items linked to historical events through \emph{causes} and \emph{enables} relationships, where $E$ is the set of connected events, $\mathrm{ImpactScore}(e)$ is the score of event $e$ from Algorithm~\ref{alg:event-impact}, and $\beta = 0.3$.
Both constants were fixed by design rather than tuned. The decay rate $\lambda = 0.1$ has only a modest effect on the current 2024--2025 corpus but lets recency matter more as the graph grows across future cycles, while the event weight $\beta = 0.3$ keeps historical events influential without letting any single event dominate, an effect the logarithmic form further limits.

\textbf{Event Impact Scoring.}
The event factor relies on pre-computed impact scores for the 250 historical events.
This event score is computed once for each historical event by Algorithm~\ref{alg:event-impact}, and it is distinct from the item score $S$ defined above. It enters $S$ only through the event factor $f_e$, so that risks and opportunities connected to more events, or to events with higher impact scores, receive a higher score.
Each event carries magnitude $M_e$, likelihood $L_e$, relevance $R_e$ (0--1), and duration $D_e$ (months).
These values belong to the curated historical event layer and are not extracted from the crawled documents.

Drawing on industry research and consultation with domain experts, we assigned each event preliminary ratings for magnitude, likelihood, and relevance, along with its duration in months, and used the LLM to refine these into the calibrated 0--1 scores that Algorithm~\ref{alg:event-impact} uses. We then select the top 100 risks and top 100 opportunities per company based on their $S$ scores for the second scoring level.

\begin{algorithm}[htbp]
\caption{Impact score computation for ontology events.}
\label{alg:event-impact}
\begin{algorithmic}[1]
\Require Set of events $\mathcal{E}$, reference year $T = 2025$
\Ensure Each event $e \in \mathcal{E}$ carries \emph{hasImpactScore}
\Statex
\For{\textbf{each} event $e \in \mathcal{E}$}
  \State $D_{\text{norm}} \gets \min(D_e / 60,\; 1)$
  \State $\text{Base} \gets (M_e + L_e + R_e + D_{\text{norm}}) / 4$
  \Statex
  \State $n_r \gets \sum_{r : e.\text{causes}(r)} r.\text{severityScore}$
  \State $n_o \gets \sum_{o : e.\text{enables}(o)} o.\text{benefitScore}$
  \State $n_t \gets \lvert\{e' : e.\text{triggers}(e')\}\rvert$
  \State $\text{Net} \gets 1 + 0.4\ln(1{+}n_r) + 0.2\ln(1{+}n_o) + 0.3\ln(1{+}n_t)$
  \Statex
  \State $\text{Decay} \gets \exp(-0.05(T - y_e))$
  \State $e.\text{impactScore} \gets \text{Base} \times \text{Net} \times \text{Decay}$
\EndFor
\end{algorithmic}
\end{algorithm}

\textbf{LLM Relevance Re-Scoring.}
The second level applies LLM-advised relevance re-scoring to adjust the domain-agnostic algorithmic ranking for the specific industry and geographic context.
For each item, GPT-4o scores its relevance to semiconductor manufacturing ($r_s$) and to the European market ($r_e$), each on a 1--10 scale.
The adjusted score is $S' = S \cdot \text{clamp}(1 + (r_s + r_e - 10)/20,\; 0.5,\; 1.5)$, which boosts highly relevant items by up to 50\% and reduces generic items by up to 50\%, while keeping the algorithmic score as the primary driver \cite{zheng2023judging}.

The final ranked lists are presented to domain experts, who provide independent rankings of the top items.
By comparing the algorithmic, LLM-adjusted, and expert rankings, we evaluate how well the automated scoring captures genuine domain importance.

\section{RESULTS}
\label{sec:results}

The pipeline produces a KG containing 76,207 risk and opportunity individuals, each scored, across the five target companies.

Before semantic clustering, the extraction stage produces 59{,}114 individuals for Infineon, 49{,}324 for Siltronic, 34{,}951 for Intel, 17{,}534 for Texas~Instruments, and 5{,}897 for Air~Liquide. Semantic clustering with $\tau{=}0.85$ and LLM-based consolidation reduces these to 17{,}567, 17{,}426, 23{,}594, 13{,}833, and 3{,}787, a decrease of 21--70\%: Infineon and Siltronic shrink the most (70.3\% and 64.7\%) owing to highly repetitive corpora, and Texas~Instruments the least (21.1\%). Intel therefore retains the most individuals (23{,}594), consistent with its broad public reporting, and Air~Liquide the fewest (3{,}787), owing to its primarily French-language disclosures.
Across all companies, 30--79\% of individuals are grouped into clusters, with the remainder passing through as singletons.
Approximately half of all clusters (45--52\%) are \emph{perfect}, meaning all members share the same taxonomy subclass, enabling deterministic consolidation; the remaining clusters have average agreement ratios between 44\% and 76\%, indicating substantial internal consistency even when subclass assignments differ.

\textbf{Category distributions.}
All 17 top-level categories are populated for every company, confirming broad taxonomy coverage.
Financial and strategic risks account for roughly half of all extracted risks. In contrast, financial and operational opportunities dominate the opportunity side, reflecting the prominence of cost optimization and process improvement themes in corporate disclosures.

\textbf{Event scoring.}
The 250 historical events receive impact scores ranging from 0.32 to 1.97 (mean~0.90).
The five highest-scoring events are all recent geopolitical or trade-policy actions (2022--2025): US export restrictions on China's chip sector (1.97), the CHIPS and Science Act (1.92), global chip export controls (1.90), Japan semiconductor export controls (1.87), and the EU Chips Act (1.85).
Temporal decay is the dominant discriminator. Events from the 2020s average 1.34 compared to 0.59 for events from the 2000s.
This dominance partly reflects our weighting scheme, since the decay rate fixes how strongly recent events are favored, and partly reflects the data, since the highest-impact events in our set are themselves recent.
The network factor, which captures an event's causal connections to risk and opportunity categories and to other events, provides differentiation among events with similar timing (mean net factor~2.80).

\textbf{Algorithmic ranking.}
Across all companies, geopolitical risk ranks first, and strategic opportunity ranks first, reflecting both the high category weight assigned to geopolitical and strategic items ($w_c{=}1.4$) and their dense connections to the historical event layer.
Opportunity scores (mean 0.97) are slightly higher than risk scores (mean 0.91).
Air~Liquide produces the lowest mean scores (0.63/0.67 for risks/opportunities). In contrast, Texas Instruments and Siltronic produce the highest (above 1.06/1.11), consistent with their narrower product portfolios generating more focused extractions that align with high-weight taxonomy categories.

\textbf{LLM relevance re-scoring.}
The second scoring layer applies semiconductor-specific ($r_s$) and European-market ($r_e$) relevance adjustments to the top-100 risks and opportunities per company.
Semiconductor scores are consistently higher than European scores, namely Intel's risks receive the highest $\bar{r}_s{=}8.20$, while Texas Instruments shows the lowest European scores ($\bar{r}_e{=}3.48$ for opportunities), consistent with its predominantly North American revenue base.
The re-scoring produces substantial reordering. Items shift by 20--33 rank positions on average, with maximum displacements up to 93 positions.
For example, Infineon's algorithmically top-ranked risk drops from rank~1 to rank~8, while a manufacturing-concentration risk rises from rank~77 to first place after receiving high relevance scores.
Intel's top opportunity shifts from generic geographic expansion (rank~1~$\to$~29) to government collaboration (rank~8~$\to$~1), reflecting the LLM's assessment that this is more directly relevant to the European semiconductor landscape.

\textbf{Domain expert validation.}
A domain expert with semiconductor industry experience ranked the top-10 risks and top-10 opportunities for each company from the pipeline's top-100 list, using the LLM-adjusted scores and item justifications as context.
Table~\ref{tab:correlation} reports Spearman's~$\rho$ between the pipeline and expert rankings both before and after the LLM re-scoring, so that the effect of the re-scoring is directly visible.

The LLM re-scoring improves alignment with expert judgment on average, most strongly for companies peripheral to the semiconductor core, although for a few companies it leaves the ranking essentially unchanged or slightly lower.
The strongest agreement occurs for Infineon risks ($\rho{=}0.97$) and Air~Liquide opportunities ($\rho{=}1.00$). At the same time, Intel shows the weakest correlations, likely because the expert prioritized execution-specific risks (e.g., Intel~18A node transition) that the pipeline's category-level weights cannot capture.
Across all companies, the average LLM-to-expert correlation is $\bar{\rho}{=}0.55$ for risks and $\bar{\rho}{=}0.72$ for opportunities, confirming that the three-layer scoring mechanism produces rankings broadly consistent with domain expertise.

\begin{table}[htbp]
\centering
\caption{Spearman $\rho$ between the pipeline and expert rankings, before (algorithmic) and after LLM re-scoring. Higher means closer to the expert.}
\label{tab:correlation}
\begin{tabular}{@{}lcccc@{}}
\toprule
\textbf{Company} & \multicolumn{2}{c}{\textbf{Risks} ($\rho$)} & \multicolumn{2}{c}{\textbf{Opportunities} ($\rho$)} \\
\cmidrule(lr){2-3}\cmidrule(lr){4-5}
& \textbf{Before} & \textbf{After} & \textbf{Before} & \textbf{After} \\
\midrule
Air Liquide & 0.49 & 0.55 & $-$0.09 & 1.00 \\
Infineon & 0.97 & 0.97 & 0.82 & 0.82 \\
Intel & 0.24 & 0.59 & 0.61 & 0.58 \\
Siltronic & 0.12 & 0.24 & 0.12 & 0.94 \\
Texas Instruments & 0.39 & 0.38 & 0.35 & 0.25 \\
\midrule
\textit{Average} & \textit{0.44} & \textit{0.55} & \textit{0.36} & \textit{0.72} \\
\bottomrule
\end{tabular}
\end{table}

\begin{figure}[ht]
\centering
\includegraphics[width=0.8\columnwidth]{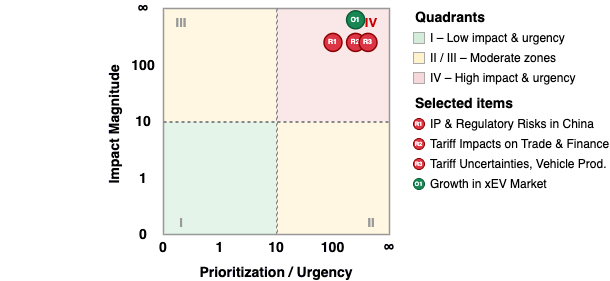}
\caption{Infineon Technologies impact-urgency matrix with selected risks (R1--R3) and opportunity (O1).}
\label{fig:matrix-infineon}
\end{figure}

\textbf{Impact-urgency matrix.}
For each company, the expert's twenty selected items are scored by the LLM on impact magnitude and prioritization urgency (both 1--10), and the three highest-priority risks and one opportunity are selected from the upper-right quadrant, illustrated for Infineon in Figure~\ref{fig:matrix-infineon}.
\emph{Trade Restrictions} appears among the selected risks for all five companies, confirming its status as a systemic cross-company threat.
The selected opportunities differ by value-chain position, including market expansion for Texas Instruments, foundry diversification for Intel, customer partnerships for Siltronic, semiconductor-adjacent growth for Air~Liquide, and automotive electrification for Infineon, indicating a complementary rather than competitive opportunity landscape for the alliance.

\section{DISCUSSION}
\label{sec:discussion}

We evaluate the pipeline along four dimensions: extraction quality, subclass assignment accuracy, clustering effectiveness, and KG utility.
We evaluate each stage with a method independent of the stage itself, and where possible a different model, so that no component grades its own output. For example, a different LLM (Claude~3.5 Sonnet) judges the extraction that GPT-4o produced.

\textbf{Extraction quality.}
An LLM-as-a-Judge protocol uses Claude~3.5 Sonnet (a different provider than the extraction model) to evaluate a stratified sample of 500~items (50~risks and 50~opportunities per company) on two criteria: \emph{discovery validity} (is the item a genuine risk or opportunity for the target company?) and \emph{subclass correctness}.
Strict precision (counting only \emph{Valid} verdicts) is 76.0\,\%; including \emph{Partial} items raises it to 92.6\,\%.
Risk extractions are more precise than opportunity extractions (85.6\,\% vs.\ 66.4\,\% strict), because risks are stated more concretely in corporate disclosures.
Critically, zero items are classified as pure hallucinations; all 37~invalid items trace to cross-company source contamination in the retrieval stage, confirming that the extraction model is faithful to its input.

\textbf{Subclass assignment accuracy.}
Overall subclass-level accuracy is 70.8\,\%, rising to 89.4\,\% at the parent-category level, indicating that GPT-4o reliably identifies the broad risk domain but struggles among semantically close siblings; multi-label subclass assignment is identified as a direction for future work.
Three recurring ambiguity patterns drive the errors: Operational vs.\ Financial (cost-causing disruptions), Technological vs.\ Strategic (technology-driven market opportunities), and Financial vs.\ Compliance (tax and subsidy items).
These reflect genuinely multifaceted phenomena rather than taxonomy defects.

\textbf{Clustering effectiveness.}
Intra-cluster cosine similarity averages 0.89 for clusters of size~$\leq$5 (85.6\,\% of all clusters), well above the $\tau{=}0.85$ threshold.
The weak Pearson correlation ($r{=}0.23$) between subclass agreement and embedding similarity reveals a design trade-off: clustering operates on short-label embeddings, while classification uses full context.

\textbf{Knowledge graph utility.}
Six competency questions validate the graph database's analytical capabilities. Key findings include: (1)~the four semiconductor manufacturers share the same top-3 risk subclasses while Air~Liquide diverges with upstream-specific risks, confirming that supply-chain cluster labels capture structural differences; (2)~30 historical events propagate across all five companies, dominated by economic and market events, validating the shared-event architecture; (3)~three-hop provenance tracing from a historical event (e.g., Global Chip Shortage~2021) to individual justified risk instances demonstrates traceability that flat reports cannot provide; (4)~80.2\,\% of risk subclasses are industry-wide (present in all five companies), confirming that the ontology captures systemic phenomena; (5)~year-over-year comparison reveals meaningful shifts, with \emph{Revenue Volatility} overtaking \emph{Regulatory Compliance} as the top risk in 2025 and \emph{Competitive Pressure} rising from rank~15 to rank~8; (6)~Jaccard similarity on top-20 risk subclasses ranges from 0.91 (Infineon--Siltronic) to 0.38 (Air~Liquide--Texas~Instruments), producing a gradient aligned with value-chain proximity.

\textbf{Scoring validation.}
The comparison (Table~\ref{tab:correlation}) shows that LLM re-scoring raises the average agreement with expert judgment from $\rho{=}0.44$ to 0.55 (risks) and from 0.36 to 0.72 (opportunities), with the largest gains for companies peripheral to the semiconductor core and little or no change for those already at its core.
For example, Air~Liquide's opportunity correlation jumps from $\rho{=}{-}0.09$ to $\rho{=}1.00$, because the relevance layer elevates semiconductor-specific items that the generic weights undervalue.
For Infineon, the algorithmic ranking alone achieves $\rho{=}0.97$ against the expert, and LLM re-scoring produces no additional improvement, suggesting that the formula's category weights are well-calibrated for companies already at the semiconductor core.
The weakest correlations occur for Intel, where the expert prioritized company-specific execution risks (e.g., the Intel~18A node transition) that receive high impact scores but are not differentiated by the pipeline's category-level weights.

\section{LIMITATIONS AND FUTURE WORK}
\label{sec:limitations}

The pipeline operates only on English-language documents, so companies like Air~Liquide, whose filings are largely in French, are under-covered. Multilingual models would directly address this gap. Source retrieval relies on 44 query templates per company executed against the Google Custom Search API, meaning that document types such as earnings-call transcripts or supplier audit reports are not covered; vision-language models for charts and infographics in annual reports could further broaden the evidence base. Cross-company source contamination, where a target company's website hosts or links to documents discussing other firms, accounts for the majority of invalid extractions; a two-stage mitigation combining a metadata pre-filter on document headers with a post-extraction attribution check would address this.
The single-linkage Union-Find clustering with a fixed threshold ($\tau{=}0.85$) can produce overly large clusters when applied to highly repetitive corpora; switching to average-linkage and conducting a sensitivity analysis across alternative thresholds would bound worst-case cluster sizes.
Clustering is performed per company, preventing ecosystem-level de-duplication of identical risks extracted independently for multiple firms; cross-company deduplication would enable a unified risk profile across the alliance. The system processes documents in a single batch run; an incremental ingestion module that monitors newly published filings and merges only new documents into the existing graph would move the system toward continuous operation. The five case-study companies populate only five of twelve ontology-defined supply chain clusters; including additional alliance-affiliated companies would strengthen cross-company findings.
Validation currently relies on a single domain expert per company. Collecting and averaging rankings from multiple experts is planned for future submissions. Finally, no formal recall measurement has been performed, as this would require ground-truth annotation by human experts; a targeted recall study on a small subset of human-annotated documents is a natural next step.

\section{CONCLUSION}
\label{sec:conclusion}

We presented a six-stage pipeline that automatically converts corporate disclosures into a scored, ontology-backed KG for semiconductor supply chain risk intelligence.
Applied to five companies spanning the ALLPROS value chain, the pipeline extracts 76{,}207 risk and opportunity individuals with 92.6\% precision, consolidates them through embedding-based clustering, and ranks them via a three-layer scoring mechanism whose LLM-adjusted outputs correlate with domain expert judgment at $\rho{=}0.55$ (risks) and $\rho{=}0.72$ (opportunities).
The resulting impact-urgency matrices identify trade restrictions as the dominant systemic risk across all value-chain positions, while the selected opportunities are complementary across companies, supporting a collaborative alliance strategy.


\footnotesize

\bibliographystyle{wsc}

\bibliography{demobib}

\section*{AUTHOR BIOGRAPHIES}

\noindent {\bf \MakeUppercase{Ema Salki\'{c}}} is a Master's student in Computer Science at TUM and a working student at Infineon Technologies AG. Her research interests include KGs, NLP, and LLM-based information extraction. Her email address is \email{ema.salkic00@gmail.com}.\\

\noindent {\bf \MakeUppercase{Alexander Fichtl}} is a Ph.D. candidate at TUM researching emergent properties of LLMs and artificial intelligence. His email address is \email{alexander.fichtl@tum.de}.\\

\noindent {\bf \MakeUppercase{Philipp Ulrich}} is an employee at Infineon Technologies AG working on semiconductor supply chain transparency using ontologies and KGs. His email address is \email{philipp.ulrich@infineon.com}.\\

\noindent {\bf \MakeUppercase{Hans Ehm}} is Senior Principal Engineer Supply Chain at Infineon Technologies AG. His research interests include semiconductor supply chains, simulation, semantic web, and machine learning. His email address is \email{hans.ehm@infineon.com}.\\

\noindent {\bf \MakeUppercase{Marta Bonik}} is a Ph.D. candidate at Infineon Technologies AG researching language technologies and the Semantic Web for supply chain knowledge management. Her email address is \email{marta.bonik@infineon.com}.\\

\noindent {\bf \MakeUppercase{Georg Groh}} is a Professor at TUM leading the Social Computing Research Group, focusing on NLP research. His email address is \email{grohg@in.tum.de}.
\end{document}